\documentclass[10pt,twocolumn,letterpaper]{article}
\usepackage{multirow}
\usepackage{siunitx}
\usepackage{makecell} 
\usepackage{bbding}
\usepackage{graphicx}
\usepackage{cuted}
\usepackage{mathtools}
\usepackage{overpic}
\usepackage{breakurl}
\usepackage[pagenumbers]{cvpr} 

\definecolor{cvprblue}{rgb}{0.21,0.49,0.74}
\usepackage[pagebackref,breaklinks,colorlinks,allcolors=cvprblue]{hyperref}

\def\paperID{} 
\def\confName{CVPR}
\def\confYear{2026}

\title{MatE: Material Extraction from Single-Image via Geometric Prior}

\author{Zeyu Zhang\textsuperscript{1} \quad
Wei Zhai\textsuperscript{1} \quad
Jian Yang\textsuperscript{1} \quad
Yang Cao\textsuperscript{1} \\[2.mm]
\textsuperscript{1}University of Science and Technology of China\\
{\tt\small \{tiptoe, yangjian12138\}@mail.ustc.edu.cn} \\
{\tt\small \{wzhai056, forrest\}@ustc.edu.cn}
}

\begin{document}
\twocolumn[{%
\renewcommand\twocolumn[1][]{#1}%
\maketitle
\vspace{-1.54em}
\includegraphics[width=1.\linewidth]{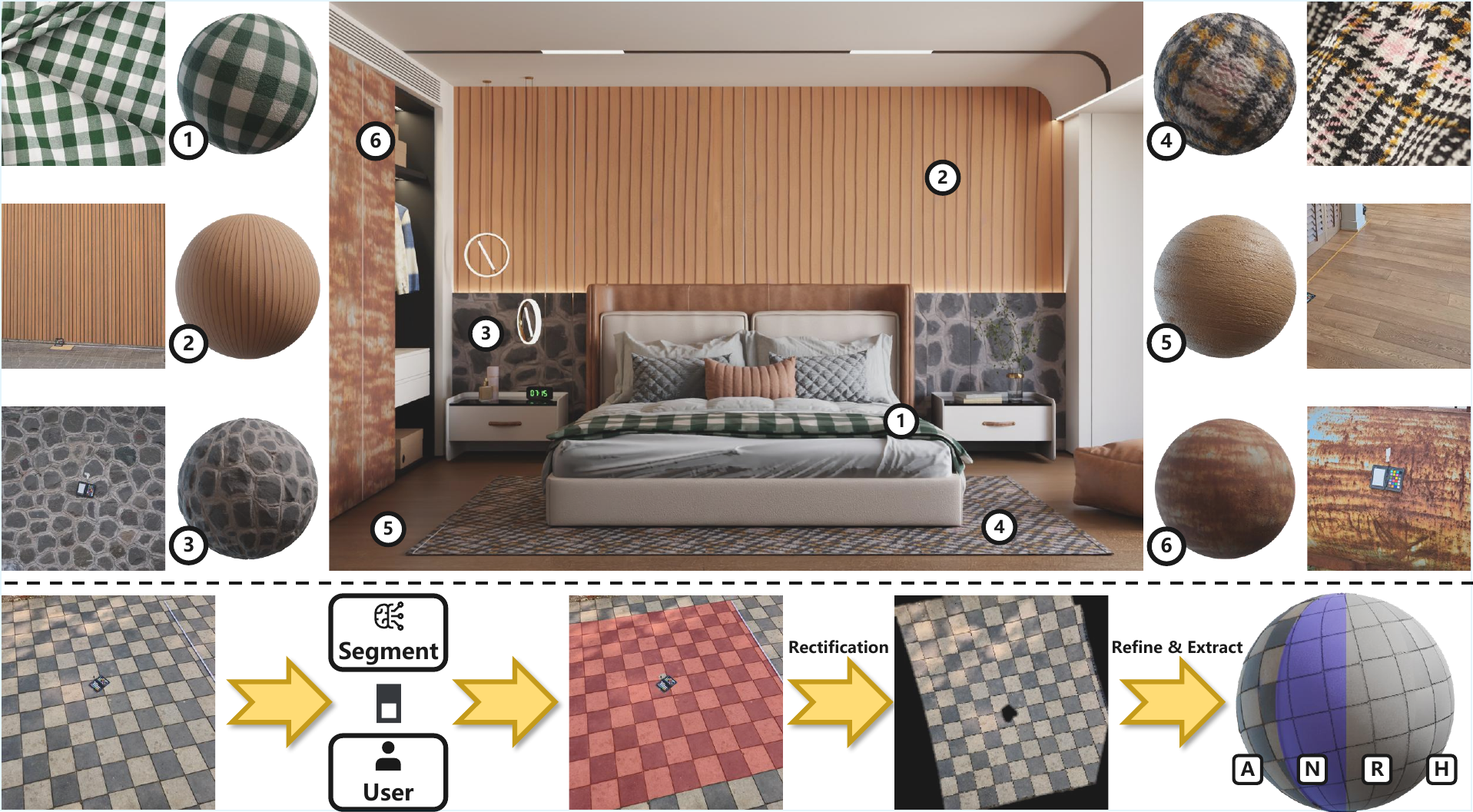}
\vspace{-1em}
\captionof{figure}{We propose MatE, a novel method for high-fidelity Physically Based Rendering (PBR) material extraction. Given a region, our method first performs rectification via geometric prior, followed by further reducing distortion and extracting the target material. The region can be sourced from user input or segmentation models like SAM~\cite{kirillov2023segment}. The extracted PBR materials(Albedo, Normal, Roughness, Height) enable the construction and texturing of realistic 3D scenes.}
\vspace{1em}
\label{fig:insight}
}]
\begin{abstract}
The creation of high-fidelity, physically-based rendering (PBR) materials remains a bottleneck in many graphics pipelines, typically requiring specialized equipment and expert-driven post-processing. To democratize this process, we present MatE, a novel method for generating tileable PBR materials from a single image taken under unconstrained, real-world conditions. Given an image and a user-provided mask, MatE first performs coarse rectification using an estimated depth map as a geometric prior, and then employs a dual-branch diffusion model. Leveraging a learned consistency from rotation-aligned and scale-aligned training data, this model further rectify residual distortions from the coarse result and translate it into a complete set of material maps, including albedo, normal, roughness and height. Our framework achieves invariance to the unknown illumination and perspective of the input image, allowing for the recovery of intrinsic material properties from casual captures. Through comprehensive experiments on both synthetic and real-world data, we demonstrate the efficacy and robustness of our approach, enabling users to create realistic materials from real-world image. \url{https://tiptoehigherz.github.io/Material-Extraction/}
\vspace{-5em}
\end{abstract}
\section{Introduction}
\label{sec:intro}
Physically-Based Rendering (PBR) materials are a cornerstone of modern computer graphics, fundamental to achieving photorealistic interactions between light and surfaces~\cite{mcauley2012practical}. However, acquiring high-quality PBR materials remains a significant challenge, often requiring specialized capture equipment and hand-crafting by expert artists~\cite{lopes2024material, asselin2020deep}. Recovering these properties from a single, in-the-wild RGB image is a particularly desirable but highly ill-posed problem~\cite{lopes2024material}. The appearance of a surface in an image is a complex entanglement of its intrinsic material properties, geometry, and unknown environmental illumination, making direct decomposition inherently ambiguous.

The advent of generative models, coupled with breakthroughs in their controllable generation~\cite{tumanyan2023plug, ye2023ip, zhang2023adding,hu2022lora} has spurred a new line of research in PBR material synthesis~\cite{guo2020materialgan, he2023text2mat, zhou2022tilegen, vecchio2024matfuse, vecchio2024stablematerials, lopes2024material, ma2025materialpicker, deschaintre2018single}. While recent efforts~\cite{lopes2024material,ma2025materialpicker} have pushed the boundaries of single-image PBR material extraction, they are nonetheless hampered by fundamental limitations. The work of Lopes et al.~\cite{lopes2024material}, for instance, learns specific texture semantics through methods like LoRA~\cite{hu2022lora} and DreamBooth~\cite{ruiz2023dreambooth}; yet, this reliance on LoRA-based adaptation inherently bakes perspective-induced distortions into the recovered material properties. Concurrently, the method proposed by Ma et al.~\cite{ma2025materialpicker} repurposes a video DiT model by conditioning it on the input image as the first frame. This paradigm, however, is predicated on a flawed premise by imposing a temporal structure on a problem that is inherently static, since the intrinsic characteristics of a material do not evolve over time. This mismatch consequently imposes a sequential dependency on the estimation, whereby minor estimation errors in early-sequence attributes are not isolated; rather, they are propagated and amplified at subsequent steps, leading to a progressive degradation in attribute fidelity (see \cref{materialpicker:sequence_problem}).

We introduce MatE, which learns a mapping from a single image to material. This bypasses the instance-specific LoRA fine-tuning of Material Palette~\cite{lopes2024material}, thus preventing overfitting to extrinsic, view-dependent properties. Furthermore, MatE employs a diffusion model for parallel material map prediction, unlike the sequential approach of MaterialPicker~\cite{ma2025materialpicker}. The significant domain misalignment between in-the-wild images and canonical PBR materials presents three distinct challenges: rectifying geometric distortion, bridging the domain gap from image to material, and preserving spatial detail. We argue that learning a direct, end-to-end mapping that implicitly handles the severe, non-linear perspective distortion is notoriously difficult. We therefore propose a coarse-to-fine strategy that decomposes this problem. First, our model employs a geometric rectification module that explicitly unwarps perspective distortion based on a geometric prior. This stage provides an initial, coarsely rectified representation, but it is not sufficient to fully resolve complex, non-linear distortions. Subsequently, we introduce a novel dual-branch diffusion architecture. It addresses the domain gap and residual distortions by leveraging the inherent image-material consistency. Furthermore, it preserves high-fidelity spatial detail by incorporating fine-grained guidance from the condition image. We propose a robust data synthesis pipeline establishing precise image-material consistency under diverse distortion and lighting. This simplifies the inverse mapping to be learned, thereby significantly improving the model's ability to bridge the synthetic-to-real domain gap.

We validate the effectiveness of MatE through extensive experiments on two challenging datasets: a synthetic dataset rendered in complex Blender scenes, and a curated real-world dataset from Polyhaven~\cite{polyhaven2025}. Our results demonstrate that MatE achieves significant improvements on the material extraction task, successfully bridging the domain gap between in-the-wild images and synthetic data. In summary, our principal contributions are as follows:

\begin{itemize}
    \item We propose MatE, a novel coarse-to-fine framework performs coarse geometry-based rectification, then extracts high-fidelity materials via a dual-branch diffusion model.
    \item We propose a robust data synthesis pipeline that generates rotation aligned image-material pairs under diverse, randomized distortion and illumination conditions.
    \item We demonstrate MatE's superior performance in reliably extracting diverse materials in complex real-world scenes.
\end{itemize}

\section{Related Works}
\label{sec:formatting}
\noindent {\bf Conditional Generative Models.} Diffusion models~\cite{songdenoising, songscore, ho2020denoising, rombach2022high} have become a leading approach in the field of generative models, demonstrating a strong capability to synthesize photorealistic images from text prompts. This success has spurred extensive research aimed at augmenting these models with more granular and diverse forms of conditional control~\cite{tumanyan2023plug, ruiz2023dreambooth, zhang2023adding, ye2023ip, hu2022lora}.

\begin{figure*}
    \centering
    \includegraphics[width=1.\linewidth]{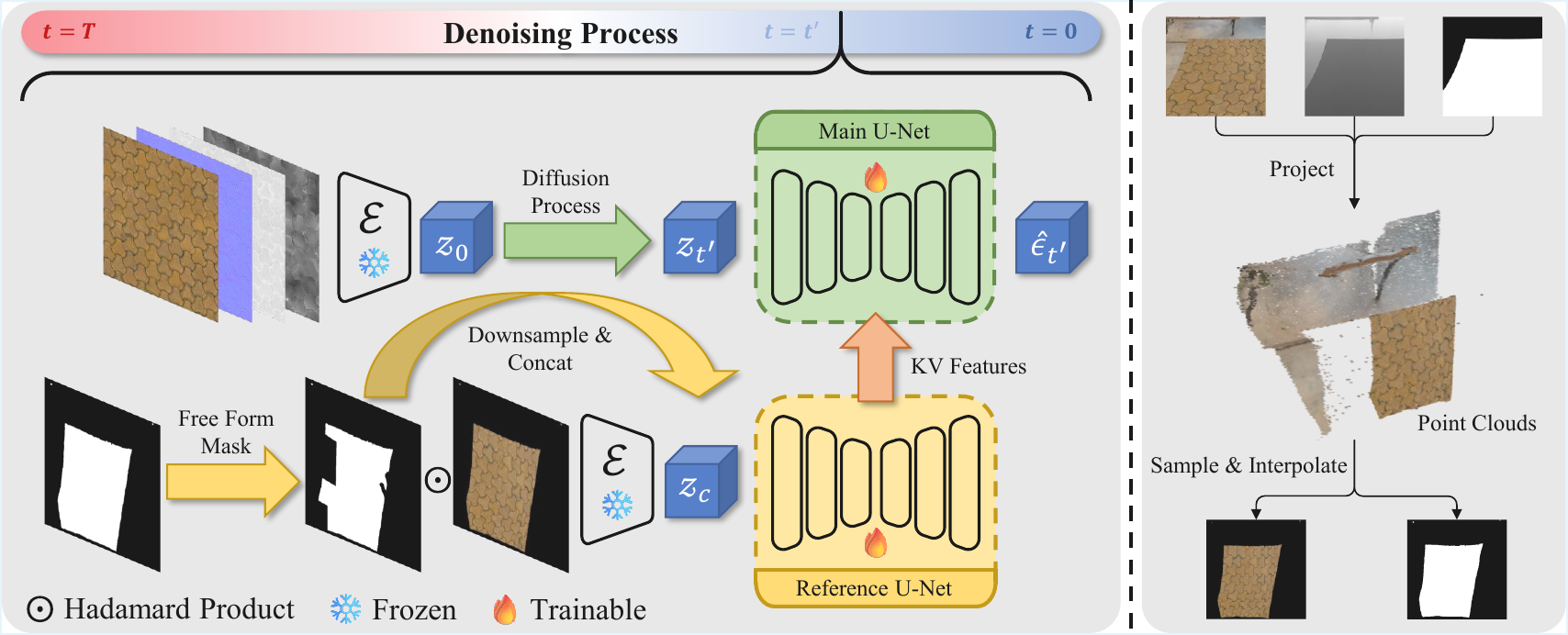}
    \label{fig:1}
    \caption{Overview of our pipeline, $\mathcal{E}$ denotes the pre-trained encoder. (\textbf{Left}) Our model consists of a Reference U-Net that processes the masked input latents to extract conditional KV features and a Main U-Net that denoises the the latent material maps ($z_{t^{\prime}}$) guided by the injected KV features. (\textbf{Right}) Visualization of our coarse rectification based on geometric prior.}
\end{figure*}
Seminal works have enabled personalization and structure-preserving editing. DreamBooth~\cite{ruiz2023dreambooth}, for instance, allows for concept injection by associating a unique token with a user-provided subject or style. Other approaches manipulate the model's internal representations during inference, such as injecting key-value pairs into cross-attention layers to guide image edits while maintaining global structure~\cite{tumanyan2023plug}. ControlNet~\cite{zhang2023adding} presents a distinct architecture, introducing a parallel, trainable branch to condition the generative process on explicit spatial inputs (e.g., edge maps), while preserving the prior of the frozen backbone. We draw inspiration from these general principles in our own framework.

\vspace{4.pt}
\noindent {\bf Texture Rectification and Generation.}
The synthesis of realistic and controllable textures is a long-standing problem in computer graphics and vision~\cite{lockerman2016multi, zhou2018non, mardani2020neural, hao2023diffusion, zhou2024generating, bellini2016time}. Conceptually, PBR material generation is strongly similar to texture generation, as both tasks share a structural need for sophisticated, spatially-aware conditioning. Early and classical approaches often rely on the principle of local self-similarity~\cite{zhai2023exploring, zhai2020deep, zhai2022exploring}, leveraging patch-based statistics to perform tasks such as texture expansion~\cite{zhou2018non, mardani2020neural}.

More recently, the advent of diffusion models has enabled a new generation of powerful and flexible methods for more complex texture manipulation tasks. For texture rectification and completion, Hao et al.~\cite{hao2023diffusion} propose a novel conditioning strategy where a deliberately distorted version of the input texture, generated via thin-plate spline transformation, is concatenated with noise and used as input to the denoising U-Net. This process guides the model to generate a complete and non-distorted output. In the domain of texture editing, Zhou et al.~\cite{zhou2024generating} achieve fine-grained control over non-stationary textures by injecting features from a reference image directly into the sampling process using a key-value (KV) injection mechanism. These works demonstrate the versatility of diffusion models in addressing diverse and challenging texture synthesis problems.

\vspace{4.pt}
\noindent {\bf Material Generation and Extraction.} 
PBR material acquisition and synthesis remain long-standing and fundamental challenges~\cite{guarnera2016brdf} in computer graphics and vision, which have witnessed significant advances in recent years driven by the development of deep learning techniques. A growing number of works have been proposed to address material intrinsics estimation~\cite{vecchio2024controlmat, zhou2022tilegen, sartor2023matfusion,kocsis2024intrinsic} and PBR material generation~\cite{guo2020materialgan, vecchio2024stablematerials, vecchio2024matfuse, he2023text2mat, guehl2020, ying2025chord, li2024procedural} from a single image or text. For instance, MatFuse~\cite{vecchio2024matfuse} achieves control over material generation using text, sketches, and images.

However, the extraction of high-fidelity PBR materials from real-world images remains an open research area. Lopes et al.~\cite{lopes2024material} propose a DreamBooth-based~\cite{ruiz2023dreambooth} technique that leverages special tokens and LoRA~\cite{hu2022lora} fine-tuning to learn and synthesize region-specific texture concepts. Ma et al.~\cite{ma2025materialpicker} extract materials by leveraging a fine-tuned video Diffusion Transformer (DiT)~\cite{peebles2023scalable} that treats condition image and different intrinsics as different frames.

\section{Method}
\subsection{Problem Statement}
\label{sec:3.1}

Our goal is to recover a set of physically-based rendering (PBR) material maps from a single, in-the-wild image. We begin by formally defining the problem.
The appearance of a real-world image $I$ results from a complex image formation process, which can be modeled by a function $f$. This function takes a set of intrinsic material properties $M$ and extrinsic environmental conditions $\mathbf{c}$ as input:

\begin{equation}
    I = f(\{ A,N,R,H \}, \mathbf{c}),
\end{equation}

\noindent where $A,N,R,H$ represent the intrinsic material properties (albedo, normal, roughness, and height). To solve this ill-posed inverse problem and recover material $\hat{M}$ from $I$, we propose a conditional diffusion model, parameterized by $\theta$, to learn this challenging inverse mapping $f^{-1}_{\theta}$. This inverse process is formulated as:

\begin{equation}
    \hat{M} = f^{-1}_{\theta}(I),
\end{equation}

\noindent where $\hat{M} = \{ \hat{A}, \hat{N}, \hat{R}, \hat{H} \}$ are the estimated material maps. Since paired ground truth data $\{ I, M \}$ from the real world is intractable, we train our model on a large-scale synthetic dataset. We generate training pairs $\{ \hat{I}, M \}$ using Blender, which serves as an approximation $\hat{f}$ of the image formation process $f$.

\begin{figure}[t]
  \centering
  \includegraphics[width=1.\linewidth]{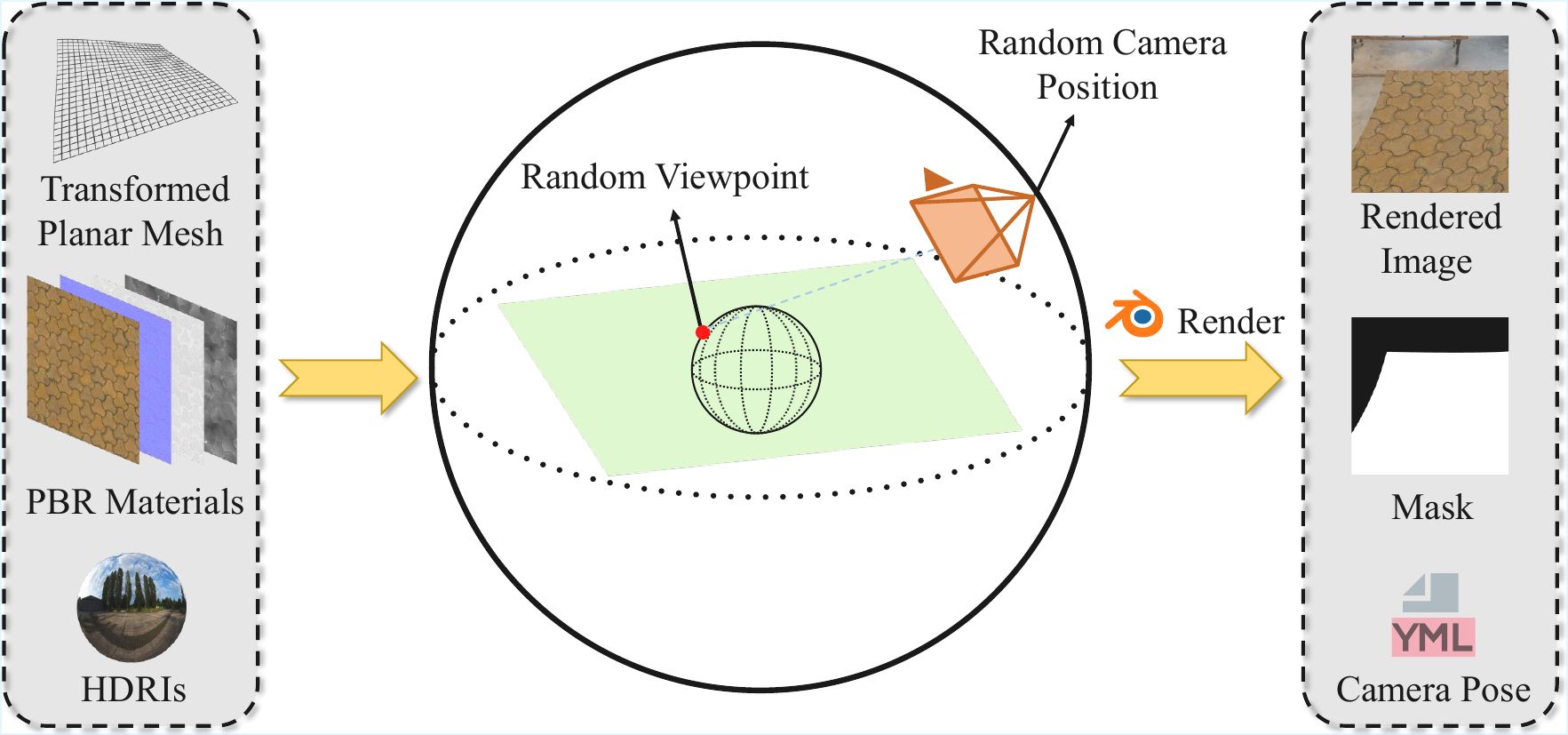}

   \caption{Overview of our dataset construction pipeline. We apply thin-plate spline (TPS) transformations to planar meshes to introduce geometric distortions. PBR materials are then mapped onto these meshes using UV coordinates, and HDRIs are employed for realistic environmental illumination. From randomly sampled camera positions and viewpoints, we then utilize Blender to render synthetic images and their corresponding masks, concurrently saving the camera poses which are essential during our training.}
   \label{fig:data_generate}
\end{figure}

\subsection{Architecture}
\label{sec:3.2}
Following the paradigm of Stable Diffusion~\cite{rombach2022high}, we first project the high-dimensional PBR material maps into a compressed latent space to ensure computational efficiency. This is achieved by individually encoding each map albedo(A), normal(N), roughness(R), and Height(H):
\begin{equation}
    \left\{ z^a, z^n, z^r, z^h \right\} = \left\{ \mathcal{E}(A), \mathcal{E}(N), \mathcal{E}(R), \mathcal{E}(H) \right\},
\end{equation}
where $z^a$, $z^n$, $z^r$, $z^h$ are the latent representations corresponding to the albedo, normal, roughness, height maps respectively and $\mathcal{E}$ denotes the encoder of a pre-trained VAE.

Our denoising network employs a dual-branch U-Net architecture to provide fine-grained conditional guidance for material extraction. Within the U-Net~\cite{ronneberger2015u} architecture of diffusion models, self-attention maps have been shown to be crucial for preserving the geometric structure and fine grained details of the input image~\cite{alaluf2024cross,zhou2024generating,tumanyan2023plug,cao2023masactrl}. We inject KV features~\cite{zhou2024generating} from the guidance images, enabling the preservation of texture details. The KV-Injection~\cite{tumanyan2023plug, cao2023masactrl} can be representated by the following equation:
\begin{equation}
    \text{Attn}(Q^{*}, K^{R}, V^{R}) = \text{Softmax}(\frac{Q^{*}({K^{R}})^{T}}{\sqrt{d}})V^{R}.
\end{equation}
Here $Q^{*} \in \mathbb{R}^{l_1 \times d}$ is the query from the main U-Net and $K^{R} \in \mathbb{R}^{l_2 \times d}$, $V^{R} \in \mathbb{R}^{l_2 \times d}$ are the key and value from the guidance image. The resulting representation has dimensions $\mathbb{R}^{l_1 \times d}$ which is agnostic to the sequence length of keys and values, thus facilitates resolution-agnostic conditioning. This process is applied across hierarchical levels of the U-Nets, enabling multi-scale feature alignment. Benifit from this design, our method is able to generate material maps at arbitrary resolutions from a single reference image with arbitrary resolution. The diffusion process of our method can be representated as following:
\begin{equation}
        \tilde{z}_t = \sqrt{\bar{\alpha}_t}z_0 + \sqrt{1-\bar{\alpha}_t}\epsilon.
\end{equation}
Here $z_0 = \mathcal{C}\left(z^a, z^n, z^r, z^d\right) \in \mathbb{R}^{b \times 16 \times h \times w}$ and $\mathcal{C}$ denotes channel-wise concatenation. We train a conditional denoising network $\epsilon_{\theta}$ to predict the applied noise $\epsilon$ from the noised latent $\tilde{z}_t$. The network is conditioned on $z_c  = \mathcal{E}\left( I \odot m \right)$. Here $I$ denotes guidance image, $m$ denotes the mask of the region of target material. The denoising network $\epsilon_{\theta}$ is then optimized using the diffusion loss objective:
\begin{equation}
    \begin{split}
        \mathcal{L}_{\text{diff}} &= \mathbb{E}_{\epsilon, t} \left[ \left\| \epsilon - \epsilon_{\theta} \left( \tilde{z}_t; z_{c}, t\right) \right\|^2 \right] \\
    \end{split}.
\end{equation}

\subsection{Perspective Projection}
\label{sec:3.3}
Learning to reverse perspective distortion in a direct, end-to-end manner is exceptionally difficult. We therefore introduce a geometric prior to decompose this challenge into a coarse-to-fine process. We first use a pre-trained depth estimation~\cite{ke2024repurposing, eftekhar2021omnidata, yang2024depth} model $\mathcal{D}$ to acquire a depth map of the input image. Based on this depth, we unproject the image and mask into a 3D point cloud and then resample it into a 2D image, yielding a coarsely rectified texture representation. This unprojection process can be formulated as:
\begin{equation}
    \left\{
    \begin{aligned}
        u_c = \mathcal{N}\bigl((u_d - c_x) \frac{\mathcal{D}(u_d, v_d) + d_{\text{shift}}}{f_x}\bigr) s_x \\
        v_c = \mathcal{N}\bigl((v_d - c_y) \frac{\mathcal{D}(u_d, v_d) + d_{\text{shift}}}{f_y}\bigr) s_y\\
    \end{aligned}
    \right.
    .
    \label{eq:project}
\end{equation}
Here, $(u_d, v_d)$ are the pixel coordinates in the distorted input image, which are mapped to the new canonical coordinates $(u_c, v_c)$. The parameters $(f_x, f_y)$ and $(c_x, c_y)$ represent the camera's intrinsic focal lengths and principal point, respectively. Since the pre-trained model $\mathcal{D}$ provides a normalized depth $\in [0, 1]$, values approaching zero would cause the unprojection to collapse to the new principal point $(c_x^{\prime}, c_y^{\prime})$. We mitigate this singularity by introducing a hyperparameter $d_{\text{shift}}$, which ensures a minimum projection distance. The core term $(u_d - c_x) \cdot (\mathcal{D}(u_d, v_d) + d_{\text{shift}}) / f_x$ first unprojects the pixel to its 3D $x$-coordinate. To ensure the resulting rectified image remains at a consistent scale independent of depth, we introduce a normalization term $\mathcal{N}$ to map the orthographic projection to the unit interval $[0, 1]$, followed by scaling factors $s_x$ and $s_y$, derived from the target image resolution, that transform these normalized values into image pixel coordinates.
\begin{figure}[t]
  \centering
  \includegraphics[width=1.\linewidth]{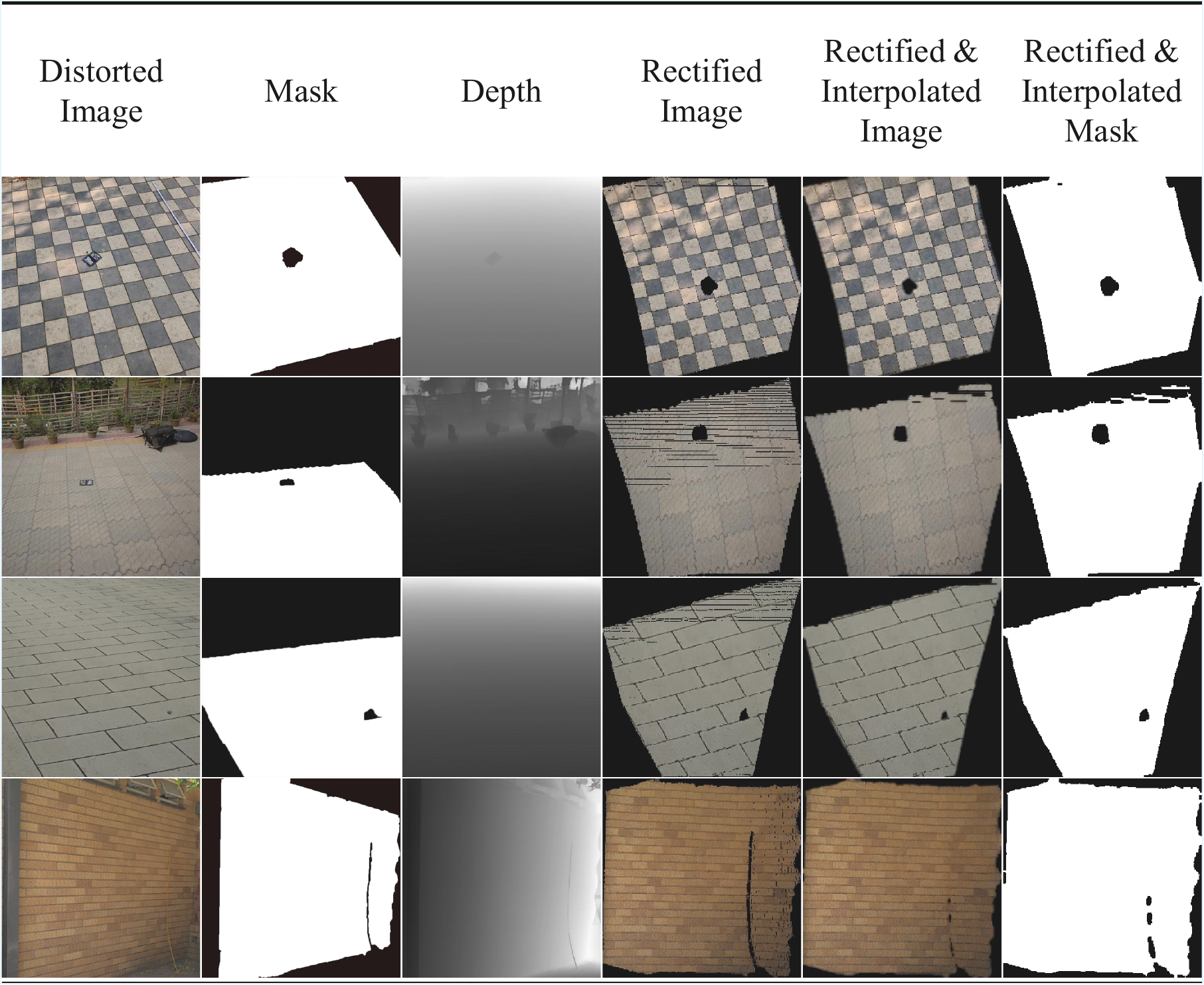}
   \caption{Our unprojection (Eq. \ref{eq:project}) generates a coarsely rectified texture, suffers from holes (Column 4). Our interpolation (Eq. \ref{eq:interpolate}) fills these artifacts to produce a dense map (Column 5).}
   \label{fig:coarse_rectify}
\end{figure}

The unprojection process described in \cref{eq:project} is a splatting operation. This process maps from a dense source grid $(u_d, v_d)$ to a sparse and irregular set of points in the canonical coordinate space $(u_c, v_c)$. When these sparse points are rasterized onto the target texture grid, this non-bijective mapping can lead to overlaps and holes. These holes are particularly prevalent in regions of texture magnification or disocclusion, degrading the quality of the coarse texture. To form a dense representation for subsequent processing, we apply a post-process interpolating operation by averaging its valid neighbors, which is defined as:
\begin{equation}
    p^{\prime}_{ij} = \left\{
    \begin{aligned}
        \sum_{m \in \mathcal{N}_k} & \sum_{n \in \mathcal{N}_k}\frac{p_{mn}}{c_s}&&, \text{if }p_{ij} \in \mathbb{H} \text{ and }c_s \neq 0 \\
        &0&&, \text{if }p_{ij} \in \mathbb{H} \text{ and }c_s = 0 \\
        &p_{ij} &&, \text{otherwise} \\
    \end{aligned}
    \right.
    ,
    \label{eq:interpolate}
\end{equation}
where $p^{\prime}_{ij}$ is the filled pixel value. $\mathbb{H}$ denotes the set of all hole pixels that received no data from the splatting. $\mathcal{N}_k(i,j)$ represents the $k \times k$ neighborhood around pixel $(i,j)$. The term $c_s$ is the total count of valid neighbors within the kernel. This operation effectively replaces each hole with the mean value of its surrounding valid pixels.

\subsection{Datasets}
\label{sec:3.4}
\noindent To enforce consistency between the view-dependent condition image and the canonical material properties, we introduce a viewpoint alignment strategy that leverages camera pose information during training. This strategy leverages camera pose information to train on rotation-aligned data, which is critical for enforcing consistency between the condition image and the material properties. Specifically, for each image in our synthetic dataset, we explicitly record the corresponding camera pose. During the training phase, this pose information is used to apply the corresponding rotation transformation directly to the canonical PBR material maps. Training a model on such view-dependent data would not require it to learn an impractically strong prior on the canonical orientation of materials.

\begin{figure}[t]
  \centering
  \includegraphics[width=1.\linewidth]{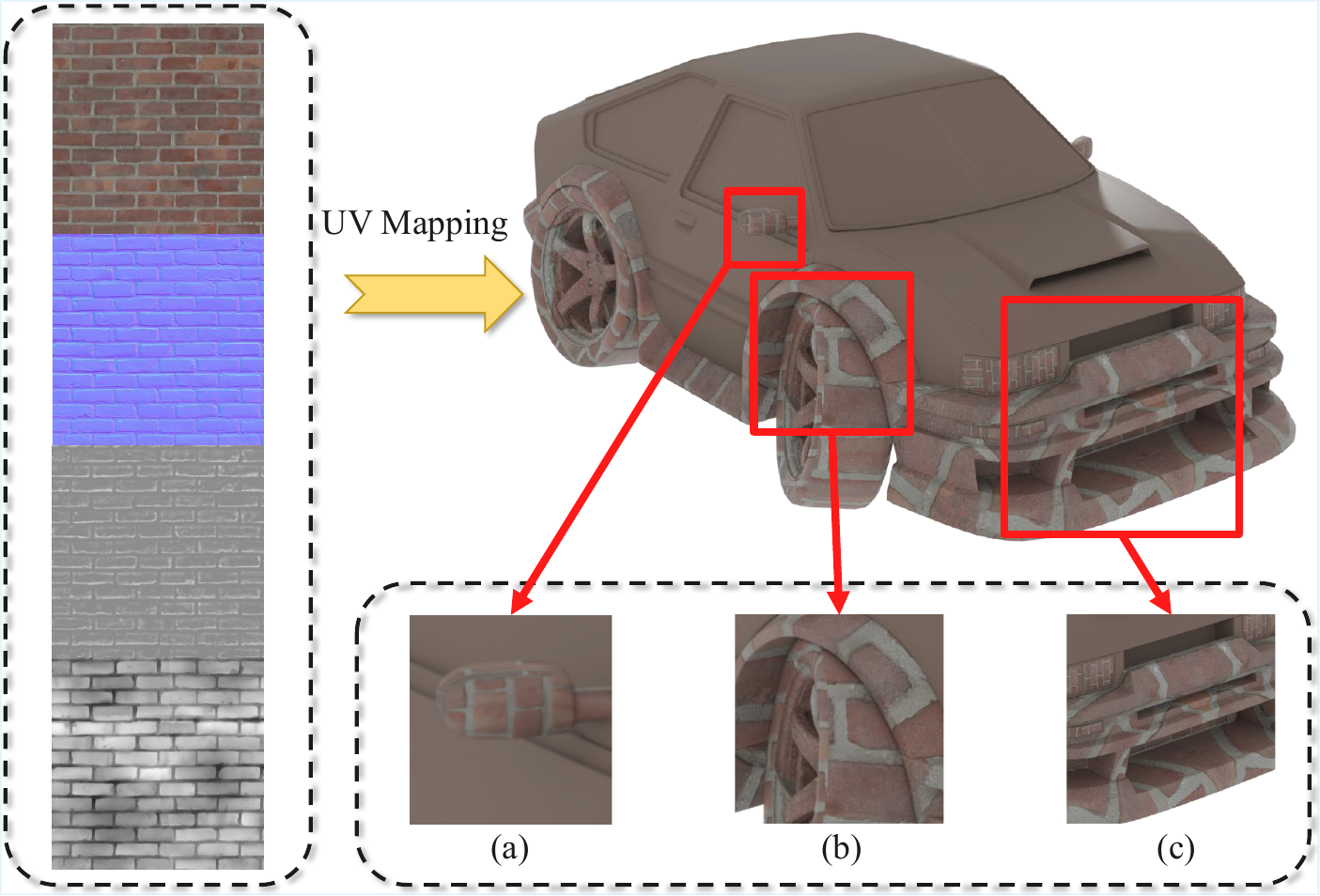}
   \caption{To circumvent artifacts such as physically implausible scaling variations(eg. inset (a)) and structural disruptions caused by discontinuous UVs(eg. insets (b) and (c)) that arise when mapping materials to complex 3D models, we generate our dataset using topologically simpler, planar meshes distorted via thin-plate spline transformation.}
   \label{fig:mesh_object}
\end{figure}

Rather than using objects from datasets like Objaverse~\cite{deitke2023objaverse}, which are prone to induce scaling variations and structural disruptions (see \cref{fig:mesh_object}), we employ a planar mesh deformed via thin-plate spline transformation~\cite{bookstein2002principal} to introduce realistic surface geometric variations. A high-fidelity material are then mapped onto this deformed surface through uv mapping\cite{enwiki:1196771315, mullen2011mastering}. Camera positions are stochastically sampled on a spherical manifold of a fixed radius. To introduce realistic viewing variations, the camera orientation is set to point towards a target point that is randomly sampled on a smaller, concentric sphere, also centered at the mesh's centroid(see \cref{fig:data_generate}). To simulate photorealistic and varied illumination, each scene is rendered under a randomly selected High Dynamic Range Image (HDRI) from Polyhaven\cite{polyhaven2025}.

\begin{figure*}
    \centering
    \begin{overpic}[width=1.\linewidth]{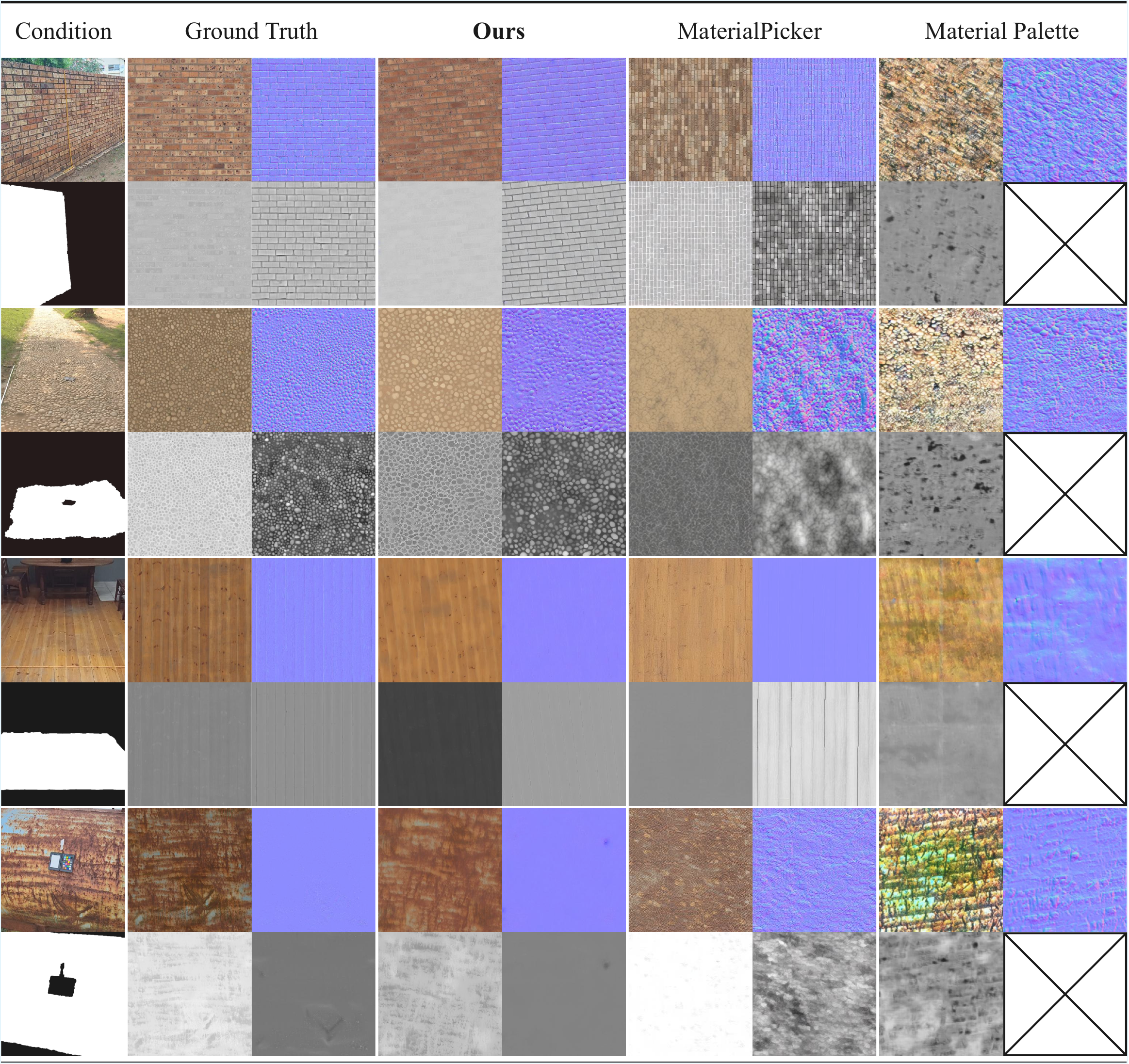} 
    \put(72.6, 91.15){$^*$}
    \put(72.9, 90.9){~\cite{ma2025materialpicker}}
    \put(95.7, 90.9){~\cite{lopes2024material}}
    \end{overpic}
    \caption{Qualitative comparison of material extraction results with state-of-the-art methods on real-world images. Within each cell, the four maps are (top-left to bottom-right) albedo, normal, roughness and height. Material Palette~\cite{lopes2024material} omits height (map left blank). $^*$ denotes our re-implementation of MaterialPicker~\cite{ma2025materialpicker}, which we built upon the Video DiT framework of~\cite{wan2025wan,wu2025improved}.} 
    \label{fig:qualitative_result}
   \vspace{-0.57em}
\end{figure*}

We define the rotation angle with camera's viewing azimuth. Formally, we define the canonical space of the planar mesh on the XY-plane, with its normal along the Z-axis. Subsequently, the alignment is achieved by an 2D rotation of the ground-truth material maps during training. The rotation angle $\alpha = \text{atan2}\left( y, x \right)$ is derived from the camera's viewing vector $\mathbf{v} = \left(x,y,z\right)$.

We render images using the PBR material dataset from~\cite{vecchio2024matsynth}, which comprises 5,879 instances. Each material instance possesses a complete set of physiologically-based attributes, including albedo, normal, roughness, and height maps. We employ the BSDF shader~\cite{mcauley2012practical} in Blender to physically accurately simulate the surface appearance of these materials.
\section{Experiment}
We conduct a comprehensive set of experiments to evaluate our proposed method. We begin by detailing the experimental setup, including the datasets, evaluation metrics, and implementation specifics in \cref{sec:4.1}. Subsequently, we present a thorough comparison of our approach against several material acquisition method~\cite{ma2025materialpicker,lopes2024material} in \cref{sec:4.2}. Finally, we investigate the key design choices regarding the model architecture, dataset construction, and training strategy, as well as the portability of our method in \cref{sec:4.3}.

\begin{table}[t] 
  \renewcommand{\multirowsetup}{\centering}
  \renewcommand{\theadalign}{cc}

  \resizebox{1.\linewidth}{!}{
  \setlength{\tabcolsep}{1pt}
  \begin{tabular}{@{} >{\centering\arraybackslash}p{2cm} c c c c c c c @{}}
      \toprule
      \multirow{2}{*}{\makecell{Method}} & \multirow{2}{*}{\makecell{Attribute}} & \multicolumn{3}{c}{\makecell{Real-World Dataset}} & \multicolumn{3}{c}{\makecell{Synthetic Dataset}} \\
      \cmidrule(lr){3-5}
      \cmidrule(lr){6-8}
      & & \makecell{LPIPS $\downarrow$} & \makecell{SSIM $\uparrow$} & \makecell{CLIP $\uparrow$} & \makecell{LPIPS $\downarrow$} & \makecell{SSIM $\uparrow$} & \makecell{CLIP $\uparrow$} \\
      \midrule
      
      \multirow{4}{=}{Material\\
      Palette~\cite{lopes2024material}} & Albedo & 0.701 & 0.214 & 0.682 & 0.529 & 0.188 & 0.720 \\
                                & Normal & 0.661 & 0.291 & 0.701 & 0.477 & 0.319 & 0.693 \\
                                & Roughness & 0.581 & \textbf{0.551} & 0.723 & 0.486 & \textbf{0.506} & 0.696 \\
                                & Height & - & - & - & - & - & - \\
      \cmidrule(r){1-8}
      
      \multirow{4}{=}{Material\\
      Picker$^{*}$~\cite{ma2025materialpicker}} & Albedo & 0.554 & 0.315 & 0.725 & 0.507 & \textbf{0.314} & 0.703 \\
                                & Normal & 0.552 & \textbf{0.317} & \textbf{0.711} & 0.408 & \textbf{0.406} & 0.693 \\
                                & Roughness & 0.455 & 0.519 & 0.751 & 0.460 & 0.437 & 0.736 \\
                                & Height & 0.542 & 0.413 & 0.686 & 0.546 & 0.338 & 0.674 \\
      \cmidrule(r){1-8}
      
      \multirow{4}{*}{\textbf{Ours}} & Albedo & \textbf{0.445} & \textbf{0.316} & \textbf{0.731} & \textbf{0.423} & 0.309 & \textbf{0.744} \\
                                & Normal & \textbf{0.395} & 0.248 & 0.704 & \textbf{0.391} & 0.303 & \textbf{0.720} \\
                                & Roughness & \textbf{0.342} & 0.509 & \textbf{0.766} & \textbf{0.345} & 0.483 & \textbf{0.753} \\
                                & Height & \textbf{0.489} & \textbf{0.437} & \textbf{0.690} & \textbf{0.508} & \textbf{0.415} & \textbf{0.683} \\
      \bottomrule
    \end{tabular}
  }
  \centering
  \caption{Quantitative comparison with state-of-the-art methods on the real-world test set collected from Polyhaven~\cite{polyhaven2025}. We report the average per-attribute LPIPS ($\downarrow$), SSIM ($\uparrow$) and CLIP-Score ($\uparrow$) between the extracted material maps and ground truth. {\bf Bold} indicates the best performance. Our method achieves state-of-the-art results across most material maps.}
  \label{tab:per_attribute_comparison_fixed}
\end{table}

\subsection{Test Datasets and Metrics}
\label{sec:4.1}
\noindent{\bf Real-World Datasets.} To evaluate the generalizability of our method, we collect 226 real-world image-material pairs publicly available from Polyhaven~\cite{polyhaven2025}. This dataset is carefully curated to include PBR materials with a wide spectrum of textures, captured under highly diverse and challenging real-world illumination. For each image, we manually annotate a precise segmentation mask, identifying the specific region of the target material.

\vspace{5.pt}
\noindent{\bf Synthetic Datasets.} While our collected real-world dataset provides a valuable benchmark for generalization, its scale is inherently limited. To facilitate a more comprehensive and large-scale evaluation, we therefore construct an additional synthetic test set. For this set, we leverage the diverse, high-quality PBR materials from Polyhaven~\cite{polyhaven2025}. We then render these materials onto three 3D scenes collected from CGTrader~\cite{cgtrader2025}, generating a new test set of 717 image-material pairs. This allows for a more extensive assessment of our model's performance across a wider spectrum of materials and environments.

\vspace{5.pt}
\noindent{\bf Evaluation Metrics.} Given that pixel-wise metrics such as MSE and PSNR are highly sensitive to translation, they are ill-suited for our task. We therefore adopt metrics that better capture perceptual and structural similarity:  LPIPS~\cite{heusel2017gans}, SSIM~\cite{wang2004image}, and a CLIP-based similarity score~\cite{radford2021learning}.

\subsection{Comparative Analysis}
\label{sec:4.2}
\noindent{\bf Quantitative Evaluation.} We quantitatively benchmark our method (MatE) against strong baselines, Material Palette and MaterialPicker, using three standard metrics: LPIPS, SSIM, and CLIP-Score. As summarized in \cref{tab:per_attribute_comparison_fixed}, our method achieves the best overall performance, significantly outperforming all baselines on key perceptual metrics LPIPS and CLIP-Score while remaining highly competitive on the SSIM. This strong performance stems from our model's unique design, which robustly handles textures degraded by viewpoint distortion and occlusions while addressing rotation differently. Whereas explicit rotation rectification would necessitate strong, often unavailable priors (regarding the canonical orientation of each texture), our model instead employs a rotation alignment mechanism.

\vspace{5.pt}
\noindent{\bf Qualitative Evaluation.} We provide visual comparisons in \cref{fig:qualitative_result}. MaterialPicker, which finetunes a video DiT model that takes different material attributes as different frames, introduces erroneous temporal dependencies. This fundamental architectural choice leads to the incorrect estimation of static material properties, as evidenced by the visual artifacts in its output. Material Palette employs a cascaded, two-stage approach: it first uses DreamBooth to generate a seamless texture and then estimates PBR materials from it. This design suffers from two critical flaws. Firstly, any viewpoint distortion present in the input image is baked into the intermediate texture, permanently degrading the final output. Secondly, as observed in our quantitative analysis, the cascaded pipeline leads to significant error accumulation, resulting in degraded PBR material quality.

\begin{figure}[t]
  \centering
  \includegraphics[width=1.0\linewidth]{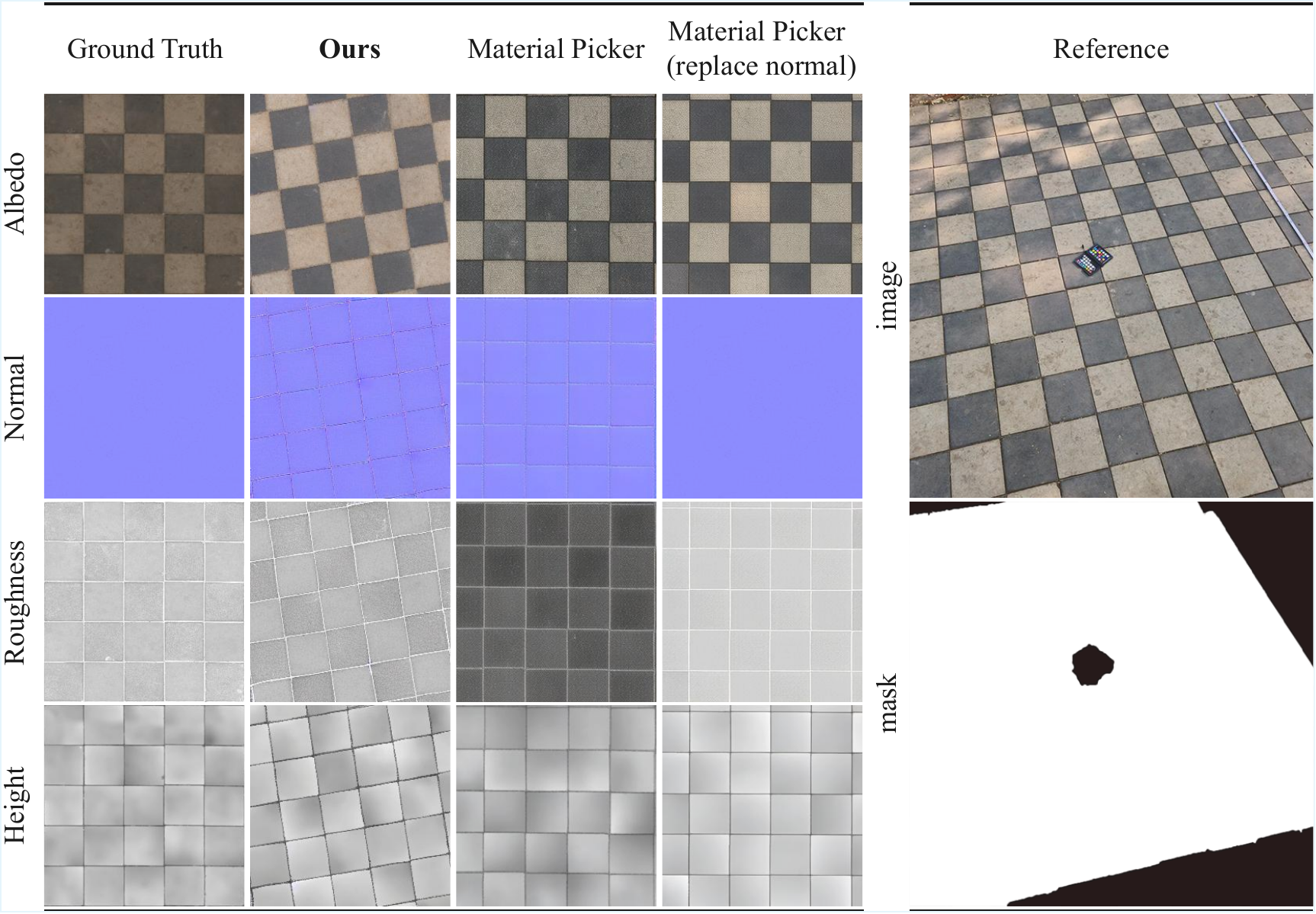}
  \caption{Qualitative ablation of cumulative error propagation. ``replace normal'': We feed the Ground-Truth normal map to MaterialPicker~\cite{ma2025materialpicker} during sampling, which restores the roughness prediction.}
  \label{materialpicker:sequence_problem}
\end{figure}

\vspace{5.pt}
\noindent{\bf Cumulative Error \& Plug-and-Play Evaluation.} We qualitatively analyze the cumulative error introduced by MaterialPicker~\cite{ma2025materialpicker} that fine-tune pre-trained video models for material generation. This approach taking material attributes as sequential video frames, imposing an autoregressive prior on a parallel task. As shown in \cref{materialpicker:sequence_problem}, we present a failure case where both our parallel model and the sequential DiT method fail to predict the normal map. Our method isolates this modular error, yielding a correct subsequent roughness map. Conversely, the finetuned Video DiT model's error propagates, causing a compounded failure in the roughness prediction. We confirm this causal link via a control experiment: providing the DiT model with the ground-truth normal map during inference restores its roughness prediction. This demonstrates the failure stems from this imposed sequential dependency, not an independent module failure, highlighting the robustness of our parallel architecture.

\begin{figure}[t]
  \centering
  \includegraphics[width=1.0\linewidth]{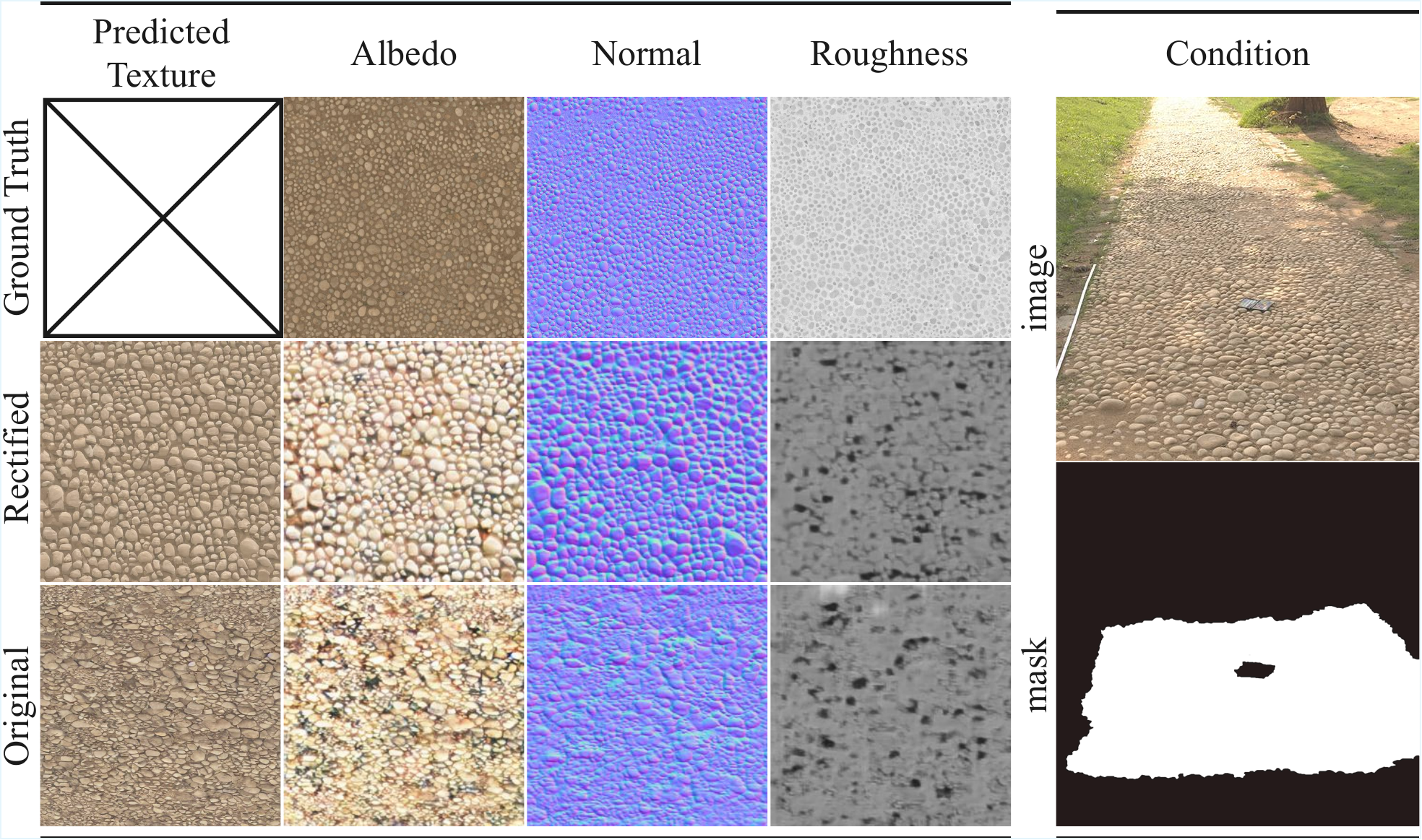}
  \caption{Qualitative results of our geometry-based coarse rectification as a plug-and-play module on Material Palette~\cite{lopes2024material}.}
  \label{figure:enhance_matpal}
\end{figure}

Our proposed geometry-based rectification method can also function as a standalone, plug-and-play module. We validate this by applying it to Material Palette \cite{lopes2024material}, which suffer from performance degradation when processing inputs with significant distortion. We perform a qualitative evaluation by applying our rectification as a pre-processing step for \cite{lopes2024material}. As shown in \cref{figure:enhance_matpal}, our module substantially boosts the baseline's performance and mitigates the degradation artifacts in its final PBR predictions.

\begin{table}[t]
  \sisetup{detect-weight, mode=text} 
  \renewcommand{\theadalign}{cc} 
  
  \resizebox{\columnwidth}{!}{%
  \setlength{\tabcolsep}{3pt}
  \begin{tabular}{@{}>{\centering\arraybackslash}p{1.5cm} c c c c c c@{}}
    \toprule
    \makecell{Method} & \makecell{Our\\ Datasets} & \makecell{Rotation\\ Aligned} & \makecell{Perspective\\ Rectified} & \makecell{LPIPS $\downarrow$} & \makecell{SSIM $\uparrow$} & \makecell{CLIP $\uparrow$} \\
    \midrule
    \multirow{4}{=}{Material\\
    Picker$^{*}$~\cite{ma2025materialpicker}}
    & \XSolidBrush & \XSolidBrush & \XSolidBrush & 0.526 & 0.391 & 0.718 \\
    & \Checkmark & \XSolidBrush & \Checkmark & 0.451 & 0.302 & 0.740 \\
    & \Checkmark & \Checkmark & \XSolidBrush & 0.452 & 0.350 & 0.732 \\
    & \Checkmark & \Checkmark & \Checkmark & 0.425 & 0.354 & \textbf{0.746} \\
    \cmidrule(r){1-7}

    \multirow{4}{*}{\textbf{Ours}} 
    & \Checkmark & \XSolidBrush & \XSolidBrush & 0.464 & 0.360 & 0.706 \\
    & \Checkmark & \XSolidBrush & \Checkmark & 0.428 & \textbf{0.396} & 0.712 \\
    & \Checkmark & \Checkmark & \XSolidBrush & 0.435 & 0.391 & 0.698 \\
    & \Checkmark & \Checkmark & \Checkmark & \textbf{0.418} & 0.377 & 0.723 \\

    \bottomrule
  \end{tabular}
  }
  \centering
  \caption{Ablation study on the effect of the rotation alignment mechanism and dataset generation method. Best results are highlighted in {\bf bold}.}
  \label{tab:ablation_checklist}
\end{table}

\subsection{Ablation Studies}
\label{sec:4.3}
\noindent{\bf Efficacy \& Portability.} We validate the efficacy of our geometry-based rectification and data generation approach via an ablation on our own MatE, and test its portability by porting it to MaterialPicker~\cite{ma2025materialpicker}. For brevity, all reported metrics are averaged across attributes (albedo, normal, roughness, and height). First, the ablation on MatE demonstrates a substantial performance gain, confirming its efficacy. Furthermore, porting our approach to MaterialPicker also yields a significant performance boost across most metrics, proving its generalizability.


\vspace{5.pt}
\noindent{\bf Condition Method.} We investigate the impact of different conditioning strategies within our framework. As detailed in \cref{tab:ablation_condition}, we benchmark our proposed method against three strong alternatives: latent concatenation, CLIP~\cite{radford2021learning} feature injection, and an addition-based mechanism inspired by ControlNet~\cite{zhang2023adding}. Our approach significantly outperforms all alternatives. While semantically rich, the high-level features from CLIP~\cite{radford2021learning} lack the fine-grained spatial details crucial for accurate material extraction, leading to noticeable distortion and loss of fidelity. Similarly, methods relying on concatenation or ControlNet~\cite{zhang2023adding} excel at tasks where the condition and the result share a high degree of spatial correlation. However, they are ill-suited for our task, which often involves significant spatial discrepancies between the input condition (the reference image) and the target output (the rectified material maps).
\label{cref:null}

\begin{table}
  \centering
  \resizebox{.8\linewidth}{!}{
  \begin{tabular}{@{} cccc }
    \toprule
    Condition Method & LPIPS $\downarrow$ & SSIM $\uparrow$ & CLIP $\uparrow$ \\
    \midrule
    CLIP~\cite{radford2021learning} & 0.674 & 0.156 & 0.649 \\
    Concat & 0.529 & 0.348 & 0.669 \\
    ControlNet~\cite{zhang2023adding} & 0.447 & 0.364 & 0.723 \\
    \textbf{Ours} & \textbf{0.418} & \textbf{0.377} & \textbf{0.723} \\
    \bottomrule
  \end{tabular}
  }
  \caption{Ablation study on conditioning mechanism.}
  \label{tab:ablation_condition}
\end{table}

\section{Conclusion}
We have introduced MatE, a novel framework that leverages geometric priors to perform rectification and extract PBR materials. While our method demonstrates robust performance in extracting textures from specified regions, we identify three primary limitations. First, MatE struggles with non-static textures that possess strong, regular internal structures. In such case our rectification process may fail to preserve the texture's structural integrity, leading to visual artifacts. Second, the coarse rectification, while generally effective, can be insufficient when faced with highly complex geometry. Third, while our approach is robust to varied illumination, it struggles with surfaces under extreme specular highlights, which can lead to extraction failures. Despite these challenges, the coarse rectification strategy employed by MatE significantly reduces the difficulty of the material extraction task, successfully bridging a critical gap between synthetic and real-world data. We believe this work provides a valuable foundation, and we hope our geometry-aware paradigm inspires future research.

\newpage
{
    \small
    \bibliographystyle{ieeenat_fullname}
    \bibliography{main}
}

\clearpage
\setcounter{page}{1}
\maketitlesupplementary

\section{More Details on Tileability}
\label{sec:rationale}
\subsection{Tileable Material Extraction}
\begin{figure}[t]
    \centering
    \includegraphics[width=1.\linewidth]{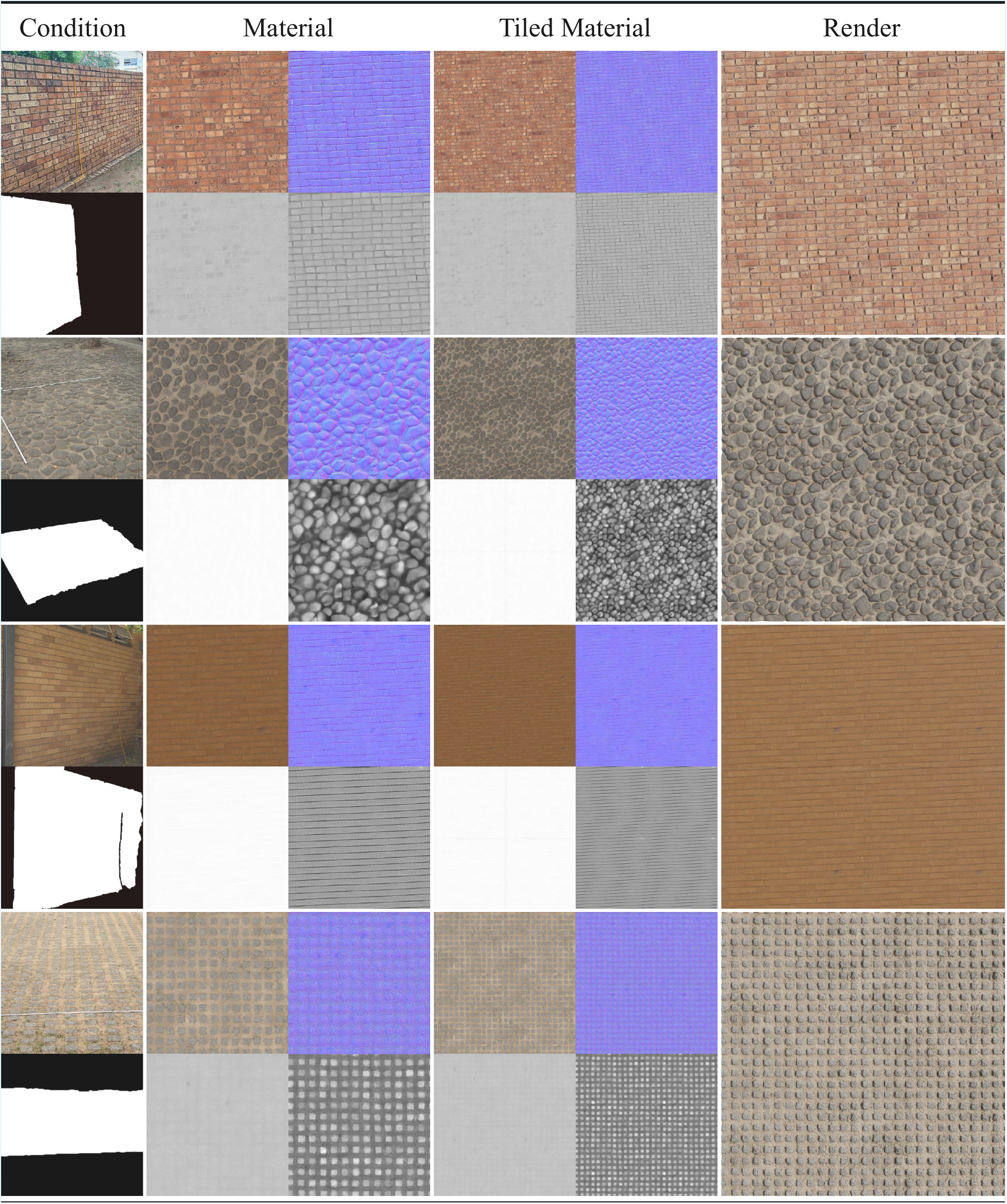}
    \caption{Visualization of generated tileable materials. The figure displays tileable materials generated by noise rolling~\cite{vecchio2024controlmat}.}
    \label{fig:tileable_generation}
    \vspace{-0.5em}
\end{figure}

Although our proposed method does not explicitly impose tileability constraints during training, we demonstrate that it can effectively generate seamless PBR materials. By incorporating the noise rolling strategy~\cite{vecchio2024controlmat} during the inference sampling stage, we achieve tileability without the need for fine-tuning. \cref{fig:tileable_generation} presents qualitative results demonstrating this capability.

\begin{figure}[b]
    \centering
    \includegraphics[width=1.\linewidth]{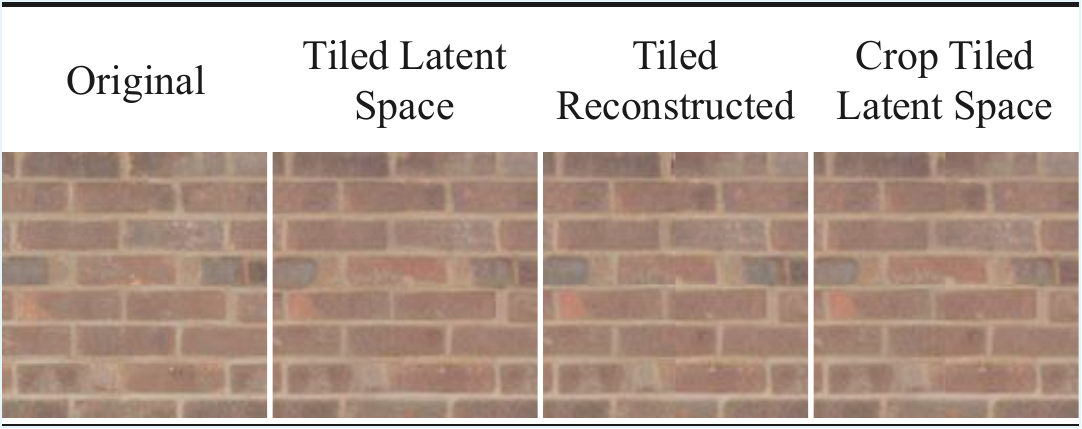}
    \caption{Comparison of tiling consistency between latent and RGB spaces.}
    \label{fig:tileable_problem}
\end{figure}

\subsection{Seam Analysis}
However, while the noise rolling strategy in ControlMat~\cite{vecchio2024controlmat} significantly facilitates boundary continuity, it does not strictly guarantee seamlessness; perceptible seams often persist when the generated materials are tiled. Furthermore, our empirical experiments reveal a notable performance discrepancy between tiling in the latent space versus tiling the final RGB outputs. We hypothesize that this variation stems from the inherent smoothing properties of the pre-trained VAE decoder, which affects how boundary discontinuities are reconstructed.

We perform a comprehensive analysis of tiling artifacts by comparing four experimental settings in \cref{fig:tileable_problem}: (1) The original tileable material (tiled in RGB space); (2) The latent-space tiled result, where latent features are tiled prior to decoding (without subsequent RGB tiling); (3) The VAE reconstruction of the original material (tiled in RGB space); (4) The top-left crop of the decoded RGB image from Setting 2, which is then tiled. As shown in \cref{fig:tileable_problem}, while the latent-space Tiled result (Col 2) is seamless, the re-tiled crop (Col 4) exhibits perceptible seams. This observation isolates the critical limitation: the VAE decoder introduces artifacts at the absolute image boundaries. While internal transitions in the latent-tiled image (Col 2) are smoothed by the decoder, the absolute boundaries are corrupted by reconstruction artifacts. This results in visible seams in the re-tiled crop (Col 4) despite the maintained structural tileability. This limitation is further evidenced by the VAE reconstruction (Col 3), where the decoder inherently introduces boundary seams into the originally seamless material.

\begin{figure*}
    \centering
    \includegraphics[width=1.\linewidth]{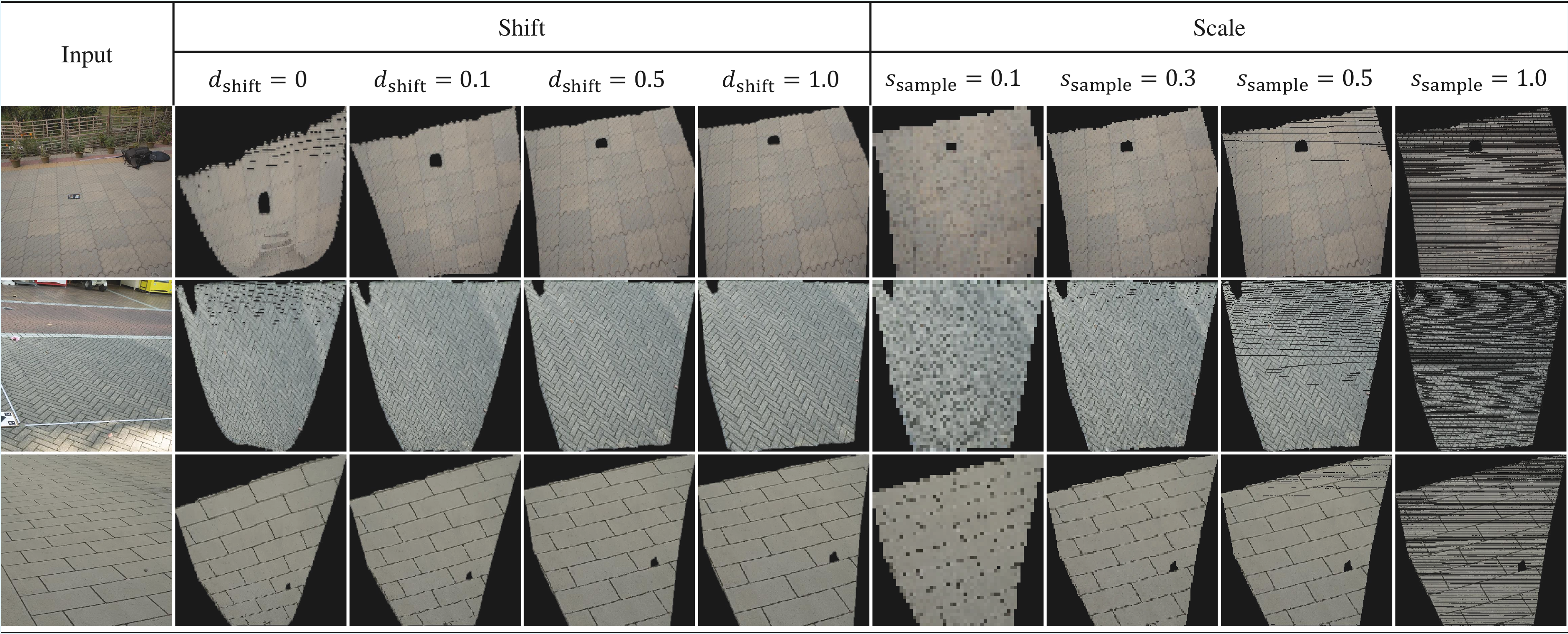}
    \caption{Sensitivity analysis of hyperparameters. We demonstrate that a large depth shift $d_{\text{shift}}$ diminishes perspective correction strength by flattening depth, while a small scale factor $s_{\text{sample}}$ fills projection holes at the cost of image sharpness.}
    \label{fig:rectification_hyperparameter}
    \vspace{-0.5em}
\end{figure*}

\section{More Details Details on Implementation}
\subsection{Dataset Implementation}
{\bf Dataset Generation.} We derive our training data from the high-quality PBR materials provided in the MatSynth dataset~\cite{vecchio2024matsynth}. To generate paired data, we utilize Blender to render corresponding images. Given the limited number of unique materials, we employ a robust data augmentation strategy to scale up the dataset. Specifically, for each material, we render 20 distinct views by randomly sampling camera poses and applying random Thin-Plate Spline (TPS) transformations to the underlying planar mesh, thereby simulating diverse geometric surface variations. To further ensure diverse illumination conditions, we utilize 100 HDRI environment maps collected from PolyHaven~\cite{polyhaven2025} during the rendering process. This combination of geometric and photometric augmentation ultimately results in a comprehensive dataset of 100k image-material pairs.

\begin{figure}[b]
    \centering
    \includegraphics[width=1.\linewidth]{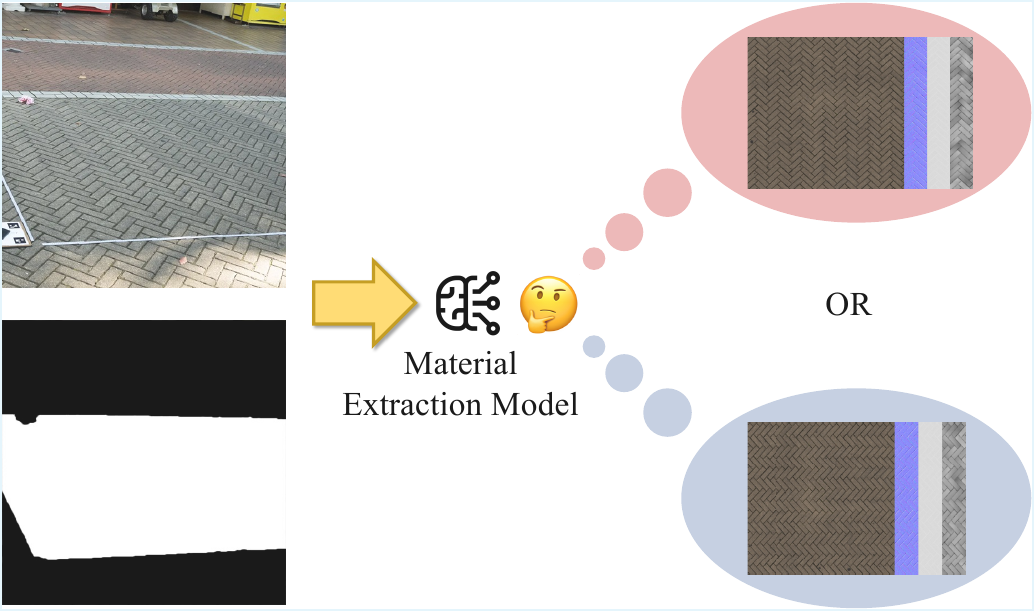}
    \caption{Illustration of orientation ambiguity. Given an input image and mask, the model will face inherent uncertainty regarding the canonical orientation of the material.}
    \label{fig:data_ambiguous}
\end{figure}

\vspace{5.pt}
{\noindent\bf Rotational Alignment Mechanism.} During the synthetic data generation pipeline in Blender, we explicitly record the camera extrinsic parameters. Utilizing this pose information allows us to bypass the need for the model to implicitly learn the canonical orientation of the materials. As illustrated in \cref{fig:data_ambiguous}, without this explicit rotational alignment, the model suffers from orientation ambiguity, being unable to uniquely determine the canonical orientation of the extracted material. By removing this degree of freedom, we simplify the learning objective, allowing the network to focus solely on material reconstruction rather than inferring the canonical orientation.

\begin{figure*}
    \centering
    \begin{overpic}[width=1.\linewidth]{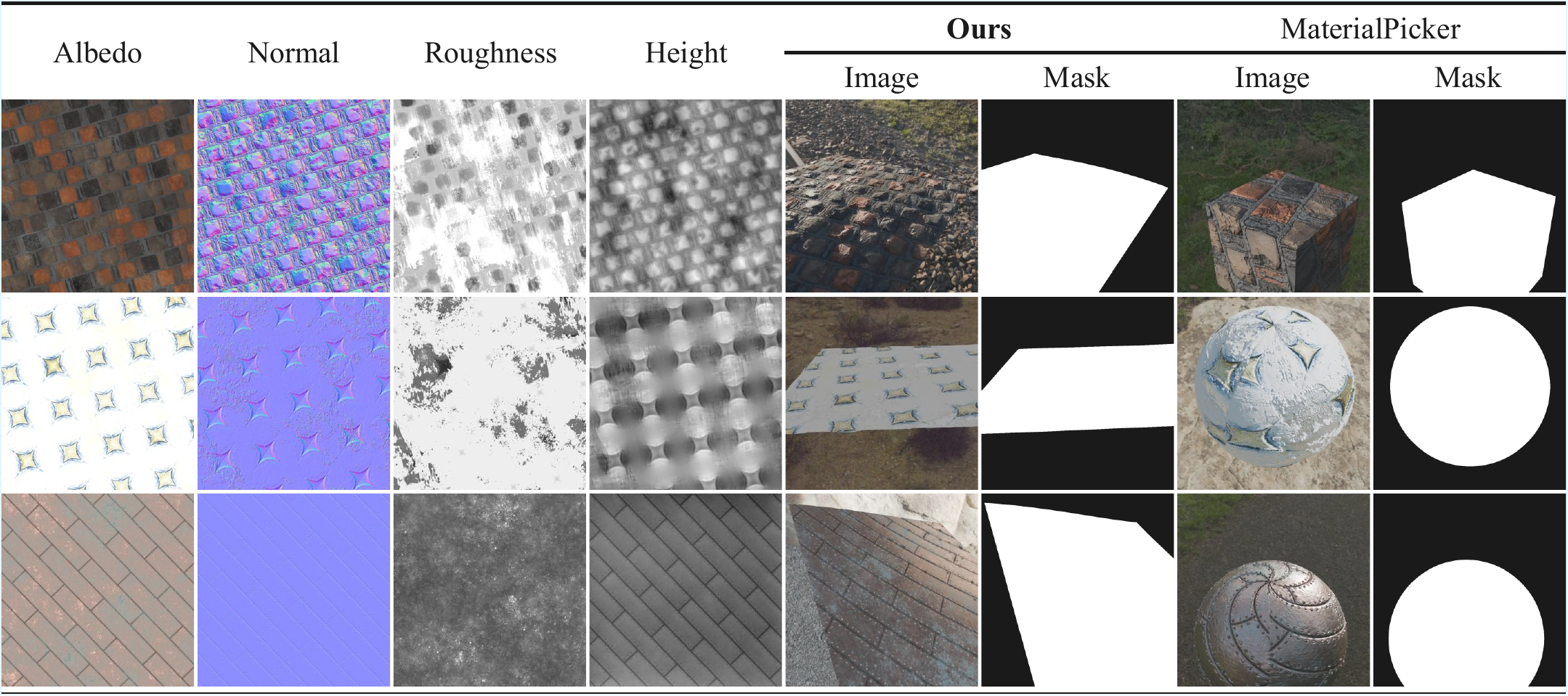} 
    \put(92.8, 42.2){\scalebox{1.}{$^*$}}
    \put(93, 42){\scalebox{.9}{~\cite{ma2025materialpicker}}}
    \end{overpic}
    \caption{Quality comparison of synthetic datasets. We compare materials generated by our pipeline against MaterialPicker~\cite{ma2025materialpicker}. MaterialPicker utilizes complex underlying geometries, which inevitably lead to severe distortions and discontinuities, breaking the structure of the texture. In contrast, our approach effectively preserves the structural integrity and coherence of the patterns.} 
    \label{fig:data_comparison}
\end{figure*}

\vspace{5.pt}
{\noindent\bf Comparison of Data Generation Pipelines.} We further compare the synthetic datasets generated by our strategy against those from MaterialPicker~\cite{ma2025materialpicker}. As visualized in \cref{fig:data_comparison}, our pipeline demonstrates superior capability in preserving the structural integrity of the textures. In contrast, MaterialPicker applies textures onto complex geometries, which inevitably introduces discontinuities and severe distortions, thereby disrupting the structural coherence of the texture patterns.


\subsection{Model Implementation}
Our method is implemented using PyTorch. We empirically observed that initializing our model with pre-trained Stable Diffusion weights did not noticeably accelerate convergence. We attribute this to the significant domain gap between the natural image distribution used in Stable Diffusion pre-training and the specific PBR material data required for our task. Consequently, we train our Denoising UNet from scratch. Our Denoising UNet follows the standard architecture of Stable Diffusion. However, to optimize GPU memory efficiency during training, we reduce channel dimensions to $[256, 512, 1024, 1024]$. Our experiments confirm that this lightweight configuration yields negligible impact on convergence performance while significantly reducing computational overhead. Regarding the dual-branch design, we removed the cross-attention modules from the Reference UNet. We extract the Key and Value features from the Reference UNet and inject them into the cross-attention layers of the Main UNet to guide the generation. For the Variational Autoencoder (VAE), we adopt the pre-trained checkpoint from Stable Diffusion v1.4 (kept frozen). We train our model using the AdamW optimizer with a fixed learning rate of $6.4 \times 10^{-5}$. The training process spans approximately $400,000$ iterations with a total batch size of 32. The entire training requires roughly 5 days on a workstation equipped with 4 NVIDIA A6000 GPUs.

\section{More Experimental Results}
\subsection{Geometry-based Rectification}
{\bf Impact of Hyperparameters.} As detailed in \cref{sec:3.3}, we introduce the hyperparameters $d_{\text{shift}}$ and $[s_x, s_y]$ to constrain the projection process. Specifically, $d_{\text{shift}}$ is designed to mitigate the ambiguity of monocular relative depth estimation. Without this shift, points with near-zero depth values would be projected erroneously close to the principal point $(c_x, c_y)$, causing severe geometric distortion. By enforcing the depth range to $[d_{\text{shift}}, 1 + d_{\text{shift}}]$, we ensure numerical stability.However, we observe a critical trade-off: excessively increasing $d_{\text{shift}}$ dampens the relative depth variations, effectively flattening the geometry and reducing the strength of the perspective correction. Consequently, a large $d_{\text{shift}}$ compromises the rectification performance by diminishing effective depth variance relative to the absolute depth. We visualize the sensitivity of our rectification results to varying $d_{\text{shift}}$ values in \cref{fig:rectification_hyperparameter}.
\begin{figure}[b]
    \vspace{-1em}
    \begin{overpic}[width=1.\linewidth]{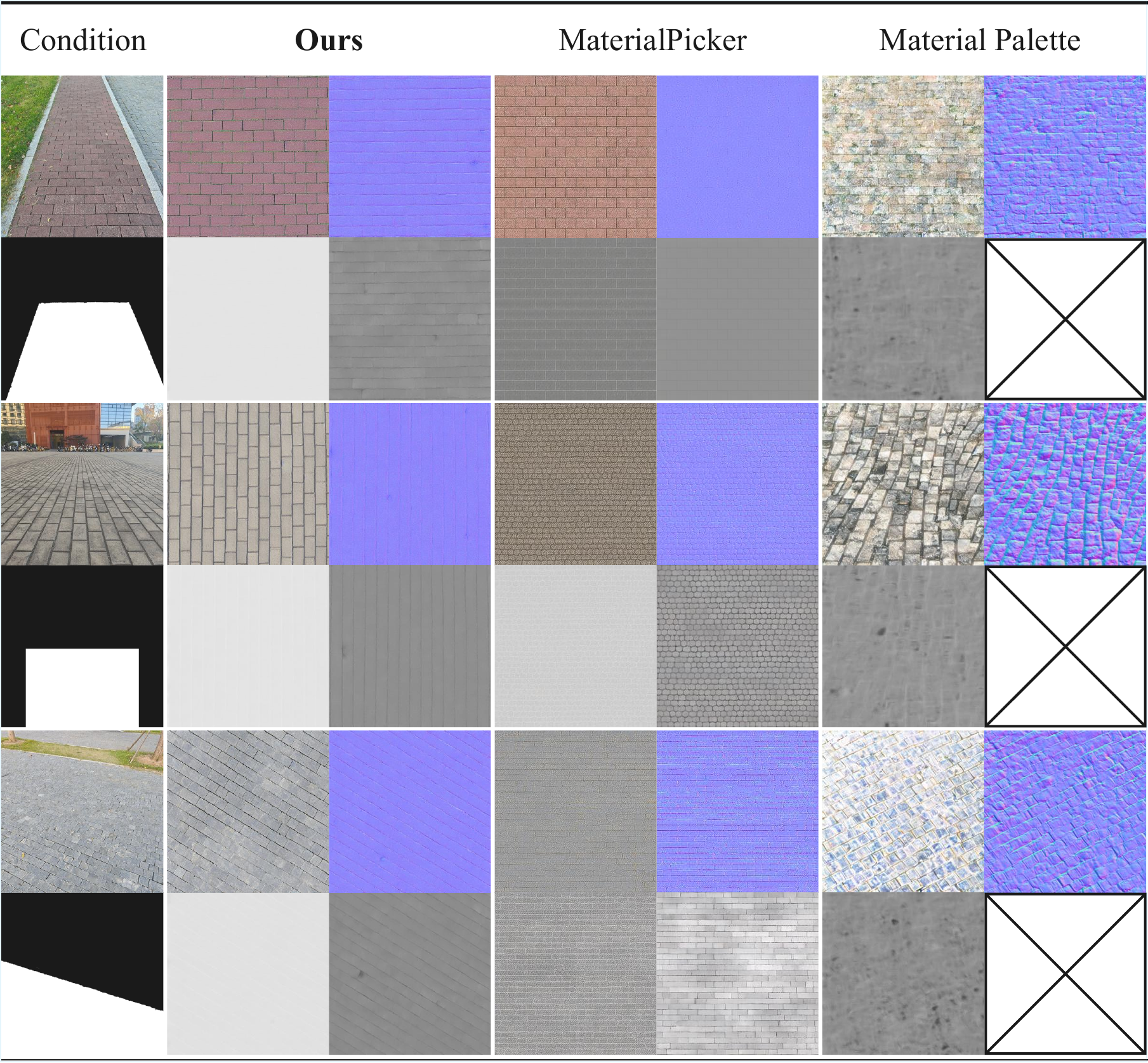} 
    \put(64.6, 88.5){\scalebox{.8}{$^*$}}
    \put(65, 88.3){\scalebox{.6}{~\cite{ma2025materialpicker}}}
    \put(94, 88.3){\scalebox{0.6}{~\cite{lopes2024material}}}
    \end{overpic}
    \caption{Qualitative results on real-world images. We present material extraction results from inputs captured by mobile devices in unconstrained environments.} 
    \label{fig:phone_capture}
\end{figure}
Simultaneously, the hyperparameter $s_{\text{sample}}$ constraints $[s_x, s_y]$ is governed by the aspect ratio changes between the source and target views. During the reprojection of the 3D point cloud back onto the 2D image plane, sparse sampling can occur, where gaps between projected points manifest as holes (empty pixels).Our experiments show that employing a smaller $s_{\text{sample}}$ effectively mitigates these artifacts by densifying the projection on a lower-resolution grid. However, to restore the output to its original resolution, we apply bilinear interpolation for upsampling. While this strategy successfully fills the holes, it inevitably introduces blurring and leads to a loss of high-frequency details. We visualize this trade-off between hole reduction and detail preservation in \cref{fig:rectification_hyperparameter}. Note that for this visualization, we intentionally omit the interpolation step defined in \cref{eq:interpolate} to clearly exhibit the raw projection results and sampling artifacts.

\begin{figure}[t]
    \centering
    \includegraphics[width=1.\linewidth]{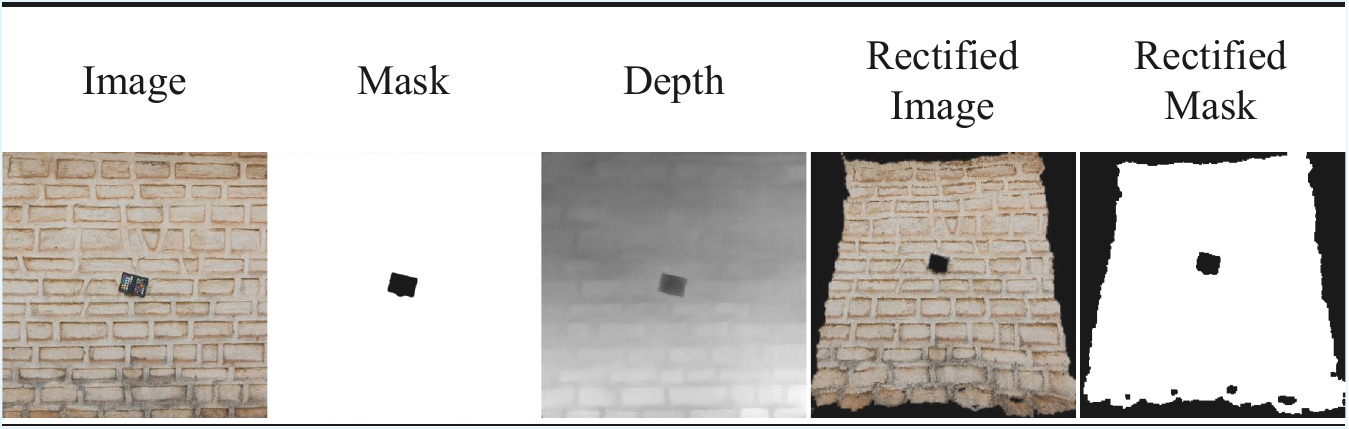}
    \caption{Limitation of geometry-based rectification on near-orthogonal views.}
    \label{fig:wrong_rectify}
    \vspace{-1.em}
\end{figure}

\vspace{5.pt}
{\noindent\bf Limitations on Near-Orthogonal Views.} One limitation arises when the camera viewpoint is near-orthogonal to the surface and the scene lacks surrounding depth cues (e.g., background objects). In such cases, the absolute depth distribution of the target surface theoretically exhibits negligible variance. However, monocular relative depth estimators inherently normalize predictions to a fixed range of $[0, 1]$. This forced normalization artificially amplifies the depth variance, introducing spurious geometric distortions into an undistorted texture, as illustrated in \cref{fig:wrong_rectify}. Despite this, the induced distortion is generally within a manageable range. Furthermore, thanks to our coarse-to-fine strategy, the subsequent generation stages are not strictly dependent on perfectly rectified inputs. Our pipeline demonstrates robustness to these minor geometric imperfections, effectively correcting or tolerating them during the refinement process.

\subsection{Real-World Examples}
To further demonstrate the generalization capability of our method, we present additional qualitative results on real-world images in \cref{fig:phone_capture}. These samples were captured using mobile devices in unconstrained environments, featuring diverse lighting conditions and viewing angles. Since ground-truth PBR maps are unavailable for such in-the-wild data, we focus on visual plausibility and quality.

\section{Discussion}
{\bf Potential of Textual Control.} While textual prompts serve as a powerful control signal in many generative tasks, we deliberately exclude them from our current pipeline. This design choice stems from the observation that high-level semantic descriptions often lack the granularity required to precisely articulate high-frequency texture details and spatial structures. Language is inherently abstract, whereas material synthesis demands pixel-level fidelity. However, we believe that integrating Vision-Language Models (VLMs) offers a promising avenue for future exploration. Specifically, leveraging the multi-turn conversational capabilities of VLMs to iteratively refine materials aligns well with user intent, allowing for precise, step-by-step adjustments to the generated output.

\vspace{5.pt}
{\noindent\bf Leveraging Pretrained Priors.} Currently, employing image-based diffusion architectures for material generation often necessitates training from scratch, a process that is both time-consuming and computationally intensive. We argue that a crucial direction for future research is to develop methodologies that can effectively leverage pretrained diffusion models directly. Unlocking the potential of these off-the-shelf priors for material synthesis would significantly reduce training costs and democratize high-quality generation.

\end{document}